\documentclass[11pt,a4paper]{article}
\usepackage[hyperref]{acl}
\usepackage{times}
\usepackage{latexsym}
\usepackage{enumitem}
\usepackage{graphicx}
\usepackage{tabularx}
\usepackage[ruled,vlined]{algorithm2e}
\usepackage{amsmath,amssymb}
\DeclareMathOperator*{\argmax}{arg\,max}
\DeclareMathOperator*{\argmin}{arg\,min}
\DeclareMathOperator{\E}{\mathbb{E}}

\usepackage{booktabs}
\usepackage{url}

\aclfinalcopy 
\def\aclpaperid{547} 

\newcommand\blfootnote[1]{%
	\begingroup
	\renewcommand\thefootnote{}\footnote{#1}%
	\addtocounter{footnote}{-1}%
	\endgroup
}

\title{Learning Dialog Policies from Weak Demonstrations}

\author{Gabriel Gordon-Hall\textsuperscript{*} \and Philip John Gorinski\textsuperscript{*}\\
	Huawei Noah's Ark Lab\\London, UK\\\texttt{gabriel.gordon.hall@huawei.com}\\\texttt{philip.john.gorinski@huawei.com}\And
	Shay B. Cohen \\
	ILCC, School of Informatics\\University of Edinburgh \\
	\texttt{scohen@inf.ed.ac.uk}\\
}

\date{}

\begin{document}
	\maketitle
	\begin{abstract}
		Deep reinforcement learning is a promising approach to training a dialog manager, but current methods struggle with the large state and action spaces of multi-domain dialog systems. Building upon Deep Q-learning from Demonstrations (DQfD), an algorithm that scores highly in difficult Atari games, we leverage dialog data to guide the agent to successfully respond to a user's requests. We make progressively fewer assumptions about the data needed, using labeled, reduced-labeled, and even unlabeled data to train expert demonstrators.
		We introduce Reinforced Fine-tune Learning, an extension to DQfD, enabling us to overcome the domain gap between the datasets and the environment.
		Experiments in a challenging multi-domain dialog system framework validate our approaches, and get high success rates even when trained on out-of-domain data.\blfootnote{*equal contribution}
	\end{abstract}
	
	\section{Introduction}
	The \emph{dialog manager} (DM) is the brain of a task-oriented dialog system. Given the information it has received or gleaned from a user, it decides how to respond. Typically, this module is composed of an extensive set of hand-crafted rules covering the decision tree of a dialog \cite{litman-plan,bos-etal-2003-dipper}. To circumvent the high development cost of writing and maintaining these rules there have been efforts to  automatically learn a dialog manager using reinforcement learning (RL; \citealt{walker2000application,Young2013POMDPBasedSS}). RL solves problems of optimal control -- where past predictions affect future states -- making it well-suited to dialog management, in which a misstep by the agent can throw the whole dialog off course. But using RL to train a dialog manager is not straightforward, and is often hindered by large dialog state spaces and sparse rewards \cite{gao2019neural}.
	
	Neural network-based deep RL \cite{mnih2015humanlevel} mitigates the problem of large state spaces \cite{fatemi2016policy,DBLP:journals/corr/LiCLG17} but it still struggles when the DM has to choose a response -- or \emph{action} -- across multiple \emph{domains} (e.g. hotel \emph{and} flight booking). In addition, deep RL performs poorly without regular feedback -- or \emph{reward} -- on the correctness of its decisions. In a dialog there is no obvious way to automatically quantify the appropriateness of each response, so RL training \emph{environments} for dialog managers usually wait until conversation-end before assigning a reward based on whether the user's task, or \emph{goal}, was completed.
	
	\begin{figure}
		\centering
		\includegraphics[width=.95\columnwidth]{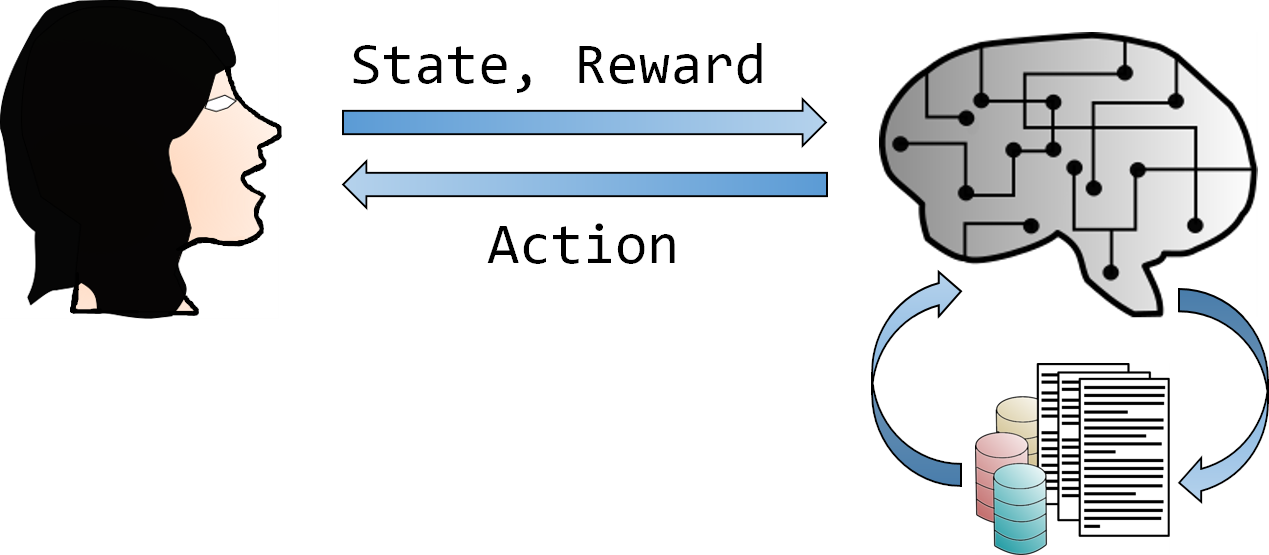}
		\caption{Illustration of reinforcement learning for dialog management. The agent (top right) interacts with the environment (left) by taking actions, and observing the resulting new state and reward. DQfD and RoFL RL agents are guided by an expert demonstrator (bottom right).}
		\label{fig:rl-diagram}
	\end{figure}
	
	An established way to deal with these difficulties is to guide the dialog manager with \emph{expert demonstrations} during RL training \cite{lipton2018bbq, us}, a high-level illustration of which is shown in Figure~\ref{fig:rl-diagram}. This approach, however, requires a rule-based oracle to provide a suitable system response given a dialog state, and does not exploit the knowledge contained in the growing number of dialog datasets \cite{budzianowski2018multiwoz,rastogi2019towards}.
	
	In this paper, we address two key-questions that arise when training RL dialog agents with expert demonstrations: (i) Can we move away from rule-based experts and use weaker, cheaper demonstrations to guide the RL dialog manager? (ii) Can we exploit information gathered during RL training to improve the demonstrator and bridge the domain gap between dialog data and the RL environment?
	
	To answer the first question, we explore three methods based on Deep Q-learning from Demonstrations  (DQfD; \citealt{hester2017}) that use trained experts derived from progressively weaker data. Our first and strongest expert is a \emph{Full Label Expert} (FLE) trained on a labeled, in-domain dataset to predict the  next system response. 
	Second, we train a \emph{Reduced Label Expert} (RLE) to predict the \emph{type} of the next system response, but not its exact nature. Finally our third expert is a \emph{No Label Expert} (NLE) that does not rely on any annotation at all, but is instead trained on unlabeled user utterance and agent response sentences. We show that all three experts can be used to successfully train RL agents, and two of them even allow us to train without expensive and often hard to come-by fully annotated in-domain dialog datasets.
	
	We address our second key question -- how to improve the experts during RL training -- by presenting \emph{\textbf{R}einf\textbf{o}rced \textbf{F}ine-tune \textbf{L}earning} (RoFL), a fine-tuning algorithm inspired by \emph{Dataset Aggregation} (DAgger; \citealt{ross2011reduction}). RoFL bridges the domain gap between dialog data and the RL environment by using the dialog transitions generated during training to update the expert's weights, adapting the previously learned knowledge to the learning environment. Our experiments show that RoFL training improves demonstrations gathered from the employed experts, giving a boost in RL performance and hastening convergence.
	
	\section{Related Work}
	Our work is closely related to research in using expert demonstrations to guide reinforcement learning dialog managers. \citet{lipton2018bbq} ``spike'' the deep Q-network (DQN; \citealt{mnih2015humanlevel}) replay buffer with a few successful demonstrations from a rule-based dialog manager. \citet{us} extend this approach and apply Deep Q-learning from Demonstrations (DQfD) to dialog, prefilling a portion of the buffer with expert transitions and encouraging the agent to imitate them by adding an auxiliary term to the DQN loss.
	
	Demonstrations are not the only way to incorporate external expertise into the dialog manager. One alternative is to use supervised learning to train a neural network policy on an in-domain dialog dataset, and then fine-tune it with policy-gradient RL on a user-simulator \cite{su2016continuously,williams2017hybrid,liu2017iterative}. \citet{liu2018dialogue} fine-tune their RL policy on human rather than simulated users. Another, parallel, approach to RL-based DMs aims to increase the frequency of meaningful rewards. \citet{takanobu2019guided} use inverse RL to learn a dense reward based on a dialog corpus, while \citet{lu2019goal} decompose the task into subgoals that can be regularly assessed.
	
	Weak demonstrations have been used outside of dialog system research to tackle RL environments with large state spaces and sparse rewards. \citet{aytar2018playing} train an expert to imitate YouTube videos of people playing challenging Atari games and exceed human-level performance. \citet{salimans2018learning} beat their score on \emph{Montezuma's Revenge} using only a single human demonstration, resetting the environment to different states from the expert trajectory. However we believe our work is the first to explore the use of weak demonstrations for DQfD in a dialog environment.
	
	RoFL, our proposed fine-tuning method, is inspired by DAgger \cite{ross2011reduction}, an iterative \emph{imitation learning} algorithm that incorporates feedback from an expert to improve the performance of a policy. DAgger requires an on-line expert that can be queried at any time, and which bounds the policy's performance. If the expert is suboptimal the policy will be too. \citet{chang2015learning} lift this restriction, allowing the policy to explore the search space around expert trajectories, but their method (LOLS) does not incorporate RL policy updates as we do.
	
	\section{Background}
	Training a dialog manager -- or \emph{agent} -- with reinforcement learning involves exposing it to an \emph{environment} that assigns a reward to each of its actions. This environment consists of a database that the DM can query, and a \emph{user-simulator} that mimics a human user trying to achieve a set of \emph{goals} by talking to the agent. The more user goals the agent satisfies, the higher its reward. Given the current state $s_t$ of the dialog, the agent chooses the next system action $a_t$ according to a policy $\pi$, $a_t = \pi(s_t)$, and receives a reward $r_t$. The expected total reward of taking an action $a$ in state $s$ with respect to $\pi$ is estimated by the Q-function:
	\begin{equation}
	Q(s, a) = \E_{\pi} \big[ \sum_{k=0}^{T-t} \gamma^{k} r_{t+k}| s_t = s, a_t = a \big]
	\end{equation}
	\begin{equation}
	\pi^*(s) = \argmax_{a \in A} Q^*(s, a)
	\end{equation}
	
	\noindent where $T$ is the maximum number of turns in the dialog, $t$ is the current turn, and $\gamma$ is a discount factor. The policy is trained to find the \emph{optimal} Q-function $Q^*(s,a)$ with which the expected total reward at each state is maximized. $\pi^*(s)$ is the optimal policy obtained by acting greedily in each state according to $Q^*$ \cite{sutton2018reinforcement}. 
	
	Deep Q-network (DQN; \citealt{mnih2015humanlevel}) approximates $Q(s, a)$ with a neural network. The agent generates dialogs by interacting with the environment, and stores
	state-action \emph{transitions} in a \emph{replay buffer} in the form ($s_t$, $a_t$, $r_t$, $s_{t+1}$). Rather than always acting according to its policy $\pi$, an $\epsilon$-greedy strategy is employed in which the agent sometimes takes a random action according to an ``exploration'' parameter $\epsilon$. Transitions aggregated in the replay buffer are sampled at regular intervals and used as training examples to update the current estimate of $Q(s, a)$ via the loss:
	\begin{equation}
	y_t = r_t + \gamma\, \max_{a'}\, Q(s_{t+1},a';\theta')
	\end{equation}
	\begin{equation}
	\mathcal{L}(Q) = (y_t - Q(s_t,a_t;\theta))^2
	\end{equation}
	
	\noindent where $\theta'$ are the fixed parameters of a \emph{target} network which are updated with the current network parameters $\theta$ every $\tau$ steps, a technique which improves the stability of DQN learning.
	
	Deep Q-learning from Demonstrations (DQfD; \citealt{hester2017}), an extension to DQN, uses expert \emph{demonstrations} to guide the agent. DQfD, prefills a portion of the replay buffer with transitions generated by the expert. The agent learns to imitate these demonstrations by augmenting $\mathcal{L}(Q)$ with an auxiliary loss term $\mathcal{L}_{aux}(Q)$:
	\begin{equation}
	\mathcal{L}_{\text{DQfD}}(Q) = \mathcal{L}(Q) + \mathcal{L}_{aux}(Q)
	\end{equation}
	The term $\mathcal{L}_{aux}$ depends on the expert used to provide demonstrations. For each of our three experts we will define a different auxiliary loss.
	
	\section{Method}
	\label{sec:method}
	It has been shown that DQfD successfully trains a dialog manager when its demonstrations come from either a rule-based, or strong pre-trained expert \cite{us}. To avoid writing rules, and to exploit the knowledge contained in external datasets, we expand on previous work and adapt DQfD for use with three progressively weaker and cheaper experts.
	Furthermore, we introduce our RoFL algorithm, describing how we fine-tune the expert during RL training.
	
	\paragraph{Full Label Expert}
	We define a \emph{Full Label Expert} (FLE) as a classifier trained on a human-to-human in-domain dialog dataset to predict, given the conversation state, the next action. For such an expert, the action space of the dataset corresponds to the actions in the RL environment and, as a result, we can use the original DQfD large margin classification term as an auxiliary loss:
	
	\begin{equation}
	\begin{aligned}
	\mathcal{L}_{aux}(Q) = & \max_{a \in A}[Q(s, a) + \ell(a_E, a)]\\
	& - Q(s, a_E)
	\end{aligned}
	\end{equation}
	
	\noindent where $a_E$ is the action the \emph{expert}
	took in $s$, and $\ell(a_E, a)$ is 0 when the agent's chosen action is the same as the action taken by the expert demonstrator, and a positive constant $c$ otherwise:
	\begin{equation}
	\ell(a_E, a)=\begin{cases}
	0, & \text{if $a = a_E$}\\
	c, & \text{otherwise}
	\end{cases}
	\end{equation}
	This FLE approach is similar to the data-driven expert introduced by \newcite{us}.
	
	\paragraph{Reduced Label Expert}
	A Full Label Expert is trained on fully-annotated in-domain data, but this is lacking for many domains, and is expensive to collect and label from scratch \cite{shah2018building}. However, although existing dialog datasets often differ in annotation, many share high-level system labels: \texttt{inform} and \texttt{request}. \texttt{inform} actions denote that the system provides information; \texttt{request} actions that the system asks for it. A system utterance from a hotel-booking dataset, e.g. ``The Le Grand Hotel costs \$48 per night, how many nights do you want to stay?'', could be labelled: [\texttt{hotel-inform-price}, \texttt{hotel-request-duration}], while a sentence from a taxi-booking dataset, e.g. ``Please let me know the dropoff location.'', could be annotated: \texttt{taxi-request-dropoff}. Although the domain and type of information are different, all actions $\mathcal{A}$ in either dataset can be broadly partitioned into sets $A_{reduced} \subset \mathcal{A}$ according to whether they \texttt{inform}, \texttt{request}, or do both.
	
\begin{figure}
		\centering
		\includegraphics[height=150px]{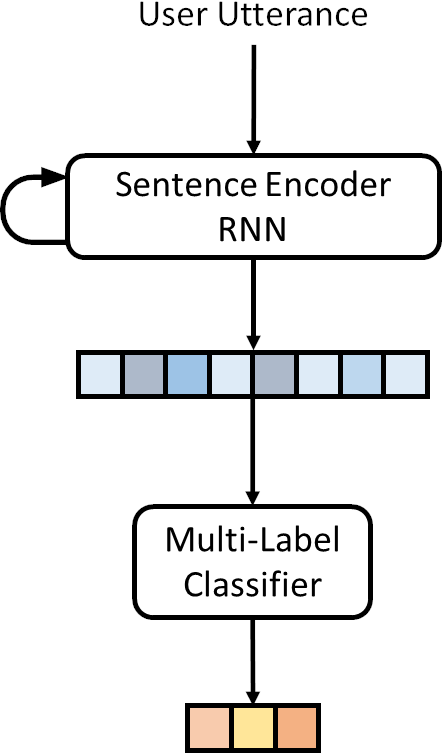}
		\caption{Reduced Label Expert (RLE) architecture.}
		\label{fig:rle}
	\end{figure}

	We introduce a \emph{Reduced Label Expert} (RLE) to take advantage of this common annotation format across diverse datasets. The RLE is a multi-label classifier that predicts the high-level annotation set $A_{reduced}$ -- or \emph{reduced} label -- of the next system action given the list $s_{NL}$ of the last few utterances in the dialog. The RLE is trained on a dialog dataset stripped down to \texttt{inform}, \texttt{request}, and \texttt{other} (for all other actions) annotations. Its architecture is outlined in Figure~\ref{fig:rle}. The previous user utterances are passed through a recurrent encoder, for example an RNN. The final hidden state of the encoder is then passed through a multi-label classifier which uses the sigmoid function to score each reduced label.%
	
	Once trained, we use the RLE to guide the dialog manager during DQfD training. First we divide all environment actions into reduced label sets. For example, the \texttt{inform} set would consist of the environment actions that pertain to providing information to the user.
	Unlike the FLE, the RLE does not predict exact actions, so we uniformly sample an environment action from the predicted reduced label set $a_E \sim A_{reduced}$ to use as an expert demonstration when prefilling the replay buffer. For example, if the RLE predicts \texttt{request} the expert might take the action \texttt{request-hotel-price}. In order to use the expert in network updates, we reformulate the $\ell$ term in the DQfD's auxiliary loss to account for the expert's reduced label prediction:
	\begin{equation}
	\ell(A_{rdcd}, s_t)=\begin{cases}
	0, & \text{if $\pi_{\theta}(s_t) \in A_{rdcd}$}\\
	c, & \text{otherwise}
	\end{cases}
	\end{equation}
	\begin{equation}
	A_{rdcd} = RLE(s_{NL})
	\end{equation}
	
	\noindent The agent is penalized by a positive constant term $c$ if the action predicted by its current policy $\pi_{\theta}$ is not in the set of actions licensed by the RLE.

	\paragraph{No Label Expert}
	While the RLE enables the use of data not annotated for the target dialog environment, it still requires labeled dialog data. This raises the question: can we employ an expert that does not rely on annotations at all?
	
\begin{figure}
		\centering
		\includegraphics[height=150px]{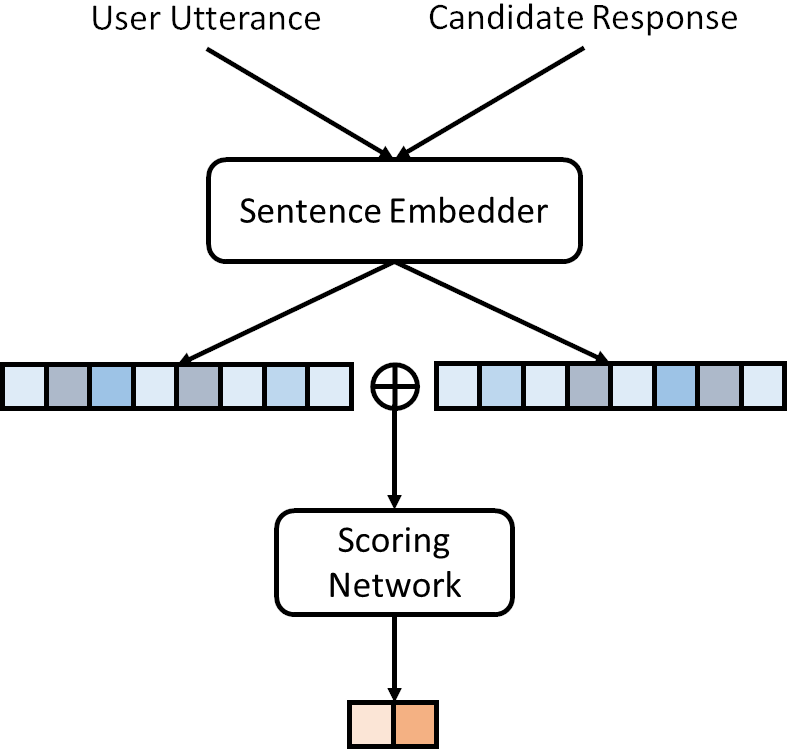}
		\caption{No Label Expert (NLE) architecture.}
		\label{fig:nle}
	\end{figure}

	To address this challenge, we propose a \emph{No Label Expert} (NLE) that uses an \emph{unannotated} dialog dataset consisting of pairs of sentences $(s_u, s_a)$, representing user utterances and the corresponding agent responses. The goal of the NLE is to predict whether, for a given pair of sentences, $s_a$ is an appropriate response to $s_u$. In this regard, it resembles models used to predict textual inference \cite{bowman2015large}. The NLE architecture is outlined in Figure~\ref{fig:nle}. The previous user utterance and a verbalized system response -- generated by an NLG component -- are consecutively passed through a sentence embedder. Their encodings are then concatenated and passed through a network which scores how appropriate the response is given the utterance.   
	
	The NLE is trained on unannotated human-to-human dialog datasets which are formatted into pairs of user utterances and agent responses. We treat these as positive instances, making the tacit assumption that in the data the agent's reply is always relevant given a user utterance. As a result, the data lacks negative examples of irrelevant agent responses. This can be mitigated by artificially creating negative pairs $(s_u, s'_a)$ from the original data by pairing each user utterance $s_u$ with random agent sentences $s'_{a}$, drawn uniformly from all agent responses that were not observed for the original $s_u$.
	Given such a dataset of positive and negative user-agent interactions, we train an NLE  that learns to output $1$ if a system response corresponds to the last user utterance, and $0$ if it does not. Once trained, we use this NLE to guide the DQfD dialog manager.
	
	When prefilling the replay buffer with expert demonstrations, we calculate the set $A_{no\, label}$ of all actions $a$ whose verbalization $s_a$ leads to an NLE output that exceeds a threshold $\rho$ when taken as a response to the last user utterance $s_u$. We then use a random action from this set $a_E \sim A_{no\, label}$ as the expert demonstration and place it in the replay buffer. We use a similar $\ell$ term in the auxiliary loss to the Reduced Label Expert, which penalizes the agent if the action $a$ predicted by its current policy is not in the set of actions licensed by the expert, i.e., if $a\not\in A_{no\, label}$:
	\begin{equation}
	\ell(A_{no\, lbl}, s_t)=\begin{cases}
	0, & \text{if $\pi_{\theta}(s_t)\in A_{no\, lbl}$}\\
	c, & \text{otherwise}
	\end{cases}
	\end{equation}
	\begin{equation}
	A_{no\, lbl}=\{a\ |\ NLE([s_u; s_a]) > \rho\}
	\end{equation}
	\noindent where $\rho$ is between 0 and 1 and $c$ is a positive constant penalty factor.
	
	\paragraph{Domain Adaptation through Fine-tuning}
	We train our experts on dialog datasets created by humans talking to humans. This data is necessarily drawn from a different distribution to the transition dynamics of an RL environment. In other words, there is a \emph{domain gap} between the two.
	
	We seek to narrow this gap by introducing \emph{\textbf{R}einf\textbf{o}rced \textbf{F}ine-tune \textbf{L}earning} (RoFL): For $d$ pre-training steps, transitions are generated according to a weak expert policy $\pi^{\xi_{\phi}}$, where the weak expert $\xi$ has parameters $\phi$. If a transition's reward exceeds a threshold $th$, we treat it as in-domain data and add it to a buffer $\mathcal{D}$. Every $\eta$ steps the expert is fine-tuned on the in-domain data gathered so far and its parameters are updated. At the end of pre-training the final fine-tuned expert's weights are frozen and its policy is used to generate demonstration transitions for another $d$ steps. This ensures that the permanent, demonstration portion of the replay buffer is filled with transitions from the fine-tuned expert. RoFL is agnostic to the expert in question and we apply it to each of our methods described above.
	
	\begin{algorithm}[h!]
		\small
		\SetAlgoLined
		\DontPrintSemicolon
		\SetKwFunction{proc}{train}
		\textbf{Inputs}: expert network $\xi$ with pre-trained parameters $\phi$, fine-tune interval $k$, a reward threshold $th$, number of pre-training steps $d$, target network update rate $\tau$, training interval $\eta$\;
		\textbf{Initialize}: random Q-network weights $\theta$, random target network weights $\theta'$, replay buffer $\mathcal{B} = \emptyset$, fine-tune data set $\mathcal{D} = \emptyset$\;\;
		\For{$t \in {1,2,...d}$} {
			Get conversational state $s_t$\;
			Sample action from expert policy $a_E \sim \pi^{\xi_{\phi}}(s_t)$\;
			Take action $a_E$ and observe $(s_{t+1},r_t)$\;
			$\mathcal{B} \gets\mathcal{B} \cup (s_t,a_E,r_t,s_{t+1})$\;
			\lIf{$r_t > th$} {
				$\mathcal{D} \gets \mathcal{D} \cup (s_t,a_E)$
			}
			\If{$t\ \textbf{\upshape mod}\ k = 0$} {
				$\phi \gets \argmin_{\phi'} -\sum_{(s, a_E)\in \mathcal{D}} a_E\log \xi_{\phi'}(s)$
			}
			\lIf{$t\ \textbf{\upshape mod}\ \eta = 0$} {
				\proc{}
			}
		}
		\For{$t \in {1,2,...}$} {
			Get conversational state $s_t$\;
			Sample action from behavior policy $a_t \sim \pi^{\epsilon Q_{\theta}}(s_t)$\;
			Take action $a_t$ and observe $(s_{t+1},r_t)$\;
			$\mathcal{B} \gets \mathcal{B} \cup (s_t,a_t,r_t,s_{t+1})$\;
			\lIf{$t\ \textbf{\upshape mod}\ \eta = 0$} {
				\proc{}
			}
		}
		\SetKwProg{myproc}{Procedure}{}{}
		\myproc{\proc{}} {
			Sample transitions from $\mathcal{B}$\;
			Calculate loss $\mathcal{L}(Q)$\;
			Perform a gradient step to update $\theta$\;
			\lIf{$t\ \textbf{\upshape mod}\ \tau = 0$} {
				$\theta' \gets \theta$
			}
		}
		\caption{Reinforced Fine-tune Learning}
		\normalsize
		\label{alg:algo1}
	\end{algorithm}
	
	\section{Experimental Setup}
	\label{sec:setup}
	We evaluate our weak experts in ConvLab \cite{lee2019convlab}, a multi-domain dialog framework based on the MultiWOZ dataset \cite{budzianowski2018multiwoz}. In ConvLab, the dialog manager's task is to help a user plan and book a trip around a city, a problem that spans multiple domains ranging from recommending attractions for sight-seeing, to booking transportation (taxi and train) and hotel accommodation.
	
	ConvLab supports RL training with an environment that includes an agenda-based user-simulator \cite{schatzmann2007agenda} and a database. The agent has a binary dialog state that encodes the task-relevant information that the environment has provided so far. This state has 392 elements yielding a state space of size $2^{392}$. In each state there are 300 actions that the DM can choose between, corresponding to different system responses when verbalized by the Natural Language Generation (NLG) module. These actions are composite and can consist of several individual informs and requests. For example, [\texttt{attraction-inform-name}, \texttt{attraction-request-area}] is one action.
	
	We train our DMs on the exact dialog-acts produced by the user-simulator, avoiding error propagation from a Natural Language Understanding (NLU) module. We use ConvLab's default template-based NLG module to verbalize system actions when using the RLE and NLE. 
	
	First, we experiment with experts trained on the \emph{in-domain} MultiWOZ dataset\footnote{We use MultiWOZ2.0 with ConvLab user annotations}. For the FLE we train on the full annotations; for the RLE we reduce the annotations to minimal \texttt{inform}, \texttt{request}, \texttt{other} labels; and for the NLE we only use the unannotated text. We also experiment with experts trained on \emph{out-of-domain} (OOD) data. To this end, we combine two datasets: Microsoft E2E \cite{li2018microsoft} -- 10,087 dialogs composed of movie, restaurant and taxi booking domains -- and Maluuba Frames \cite{el-asri-etal-2017-frames} which is made up of 1,369 dialogs from the flight and hotel booking domains. While three of these domains are also in MultiWOZ, the specifics of the conversations are different.
	
	Our Full Label Expert is a feedforward neural network (FFN) with one 150 dimensional hidden layer, ReLU activation function and 0.1 dropout which takes the current dialog state as input. The Reduced Label Expert uses the last utterance in the conversation as context, which is embedded with 300 dimensional pre-trained GloVe embeddings \cite{pennington2014glove}, then passed through a uni-directional 128 dimensional hidden layer GRU \cite{cho2014properties} from which the last hidden state is used to make a multi-label prediction. Finally, our No Label Expert uses pre-trained $\text{BERT}_{\text{base-uncased}}$\cite{devlin2018bert} to embed and concatenate user and agent utterances into 1536-dimensional input vectors, and employs a feedforward neural network with SELU activations \cite{klambauer2017self} to predict whether the agent's response is an appropriate answer to the last user utterance. Note that the RLE and NLE both take natural language as input yet use different word embeddings. We conducted preliminary experiments to evaluate the efficacy of BERT and GloVe embeddings for the respective expert training tasks. While we found that the NLE greatly benefited from BERT over GloVe, the RLE performance did not differ between embeddings. Since GloVe vectors yield a significant runtime advantage over the course of RL training, we used GloVe for the RLE, while employing slower BERT embeddings for the NLE due to the significantly better performance.
	
	For RL training of our DQfD agents, we use a prioritized replay buffer \cite{schaul2015prioritized} with a maximum buffer size of 100,000 transitions. We follow the DQfD setup of \cite{us} and apply L2 regularization with a weight of $10^{-5}$ and drop the n-step term from the original DQfD loss. All RL networks have a 100 dimensional hidden layer, a dueling network structure, and use the double DQN loss \cite{wang2015dueling,van2016deep}. All our networks are trained with the RAdam optimizer \cite{liu2019variance} with a learning rate of 0.01. For a complete list of hyperparameters used for our experiments refer to the attached Supplemental Material.
	
	We slightly alter the RoFL algorithm presented in 
	\ref{sec:method}
	to account for the fact that ConvLab only rewards the agent based on whether it successfully completed the task at the end of a dialog (intermediate steps are uniformly assigned a -1 step penalty). Rather than immediately adding transitions to the fine-tune dataset $\mathcal{D}$, we wait until the end of a conversation and check if its total reward exceeds a threshold $th$. If it does, we assume that all transitions in that conversation are perfect, and add them to $\mathcal{D}$. For our experiments we empirically determine $th$, and set it to 70.
	
	We train all our RL-based dialog managers for 3 sessions of 2,500,000 steps, and anneal the exploration parameter $\epsilon$ over the first 500,000 to a final value of 0.01. Results and training graphs in the following section are the average of these 3 sessions. Each session takes under 10 hours on one NVIDIA GeForce RTX 2080 GPU. We compare our approach to supervised and reinforcement learning baselines.
	
	\section{Results}
	Table~\ref{tab:baselines} shows evaluation results over 1,000 dialogs for baseline and DQfD dialog managers using our three proposed experts inside ConvLab's evaluation environment. The Rule baseline is a rule-based DM included in ConvLab. FFN is a supervised learning baseline DM that directly uses the same in-domain classifier introduced in Section~\ref{sec:method} to predict the next action. It is trained on MultiWOZ, and achieves 21.53\% accuracy on the test set. Deep Q-network (DQN) is an RL agent which uses the hyperparameters described in Section~\ref{sec:setup} except that it does not use demonstrations. We also compare against an agent trained with Proximal Policy Optimization (PPO; \citealt{ppo}), an actor-critic based RL algorithm widely used across domains. We use the PPO hyperparameters laid out in \citet{takanobu2019guided}. The middle third of Table~\ref{tab:baselines} summarizes results for DQfD agents trained with rule-based (RE), Full Label (FLE), Reduced Label (RLE), and No Label (NLE) experts.
	The bottom third shows results for our weak expert methods trained with RoFL (+R). We follow \citet{takanobu2019guided} and report evaluation results in terms of average dialog length (Turns), F1-Score of the information provided that was requested by the user, Match Rate of user-goals, and Success Rate -- the percentage of dialogs in which all information has been provided and all booking information is correct.
	
	\begin{table}[t]
		\centering
		\begin{tabular}{l|c|c|c|c}
			\toprule
			& Turns & Inform & Match & Success \\
			\midrule
			Rule & 5.25 & 94.00 & 100 & 100\\ 
			FFN & 11.67 & 81.00 & 52.63 & 61.00\\
			DQN & 18.79 & 28.50 & 11.07 & 11.85\\
			PPO & 5.79 & 65.67 & 72.51 & 63.27 \\
			\midrule
			RE & 5.33 & 92.33 & 97.07 & 98.33\\
			FLE & 6.81 & 89.67 & 94.12 & 91.67\\
			RLE & 7.64 & 81.33 & 89.34 & 85.03\\
			NLE & 7.20 & 84.67 & 85.31 & 86.83 \\
			\midrule
			FFN-ft & 9.62 & 83.00 & 90.79 & 76.00\\
			FLE+R & 6.75 & \textbf{90.00} & \textbf{94.57} & 92.47\\
			RLE+R & \textbf{6.38} & 88.67 & 90.62 & \textbf{92.93}\\
			NLE+R & 6.89 & 89.00 & 92.68 & 91.00\\
			\bottomrule
		\end{tabular}
		\caption{Evaluation results of baseline systems (top) as well as DQfD with rule-based and our weak expert approaches trained \emph{in-domain}. The middle section denotes DQfD agents trained without RoFL; the bottom section shows results for agents trained with RoFL. Evaluation is conducted using an agenda-based user-simulator for 1000 dialogs. Reported scores are average number of Turns, Inform F1, Match Rate, and Success Rate. Best performing \emph{weak} expert agents are in bold.}
		\label{tab:baselines}
	\end{table}
	
	As expected, the Rule agent -- written specifically for ConvLab -- almost perfectly satisfies user goals. FFN is considerably worse, with a 40\% lower Success Rate, and half the Match Rate of the rule-based agent. For standard DQN, the environment's large state and action spaces pose a serious challenge, and it barely exceeds 11\% Success and Match Rates. PPO achieves a respectable 63\% success rate, outperforming the FFN baseline.
	Crucially, \emph{all} DQfD agents significantly out-perform the FFN, DQN, and PPO baselines, with the RE and FLE approaches coming within 3\% and 6\% respectively of the Rule agent's performance.
	
	In the remainder of this section we will further analyze and compare the performances of DQfD agents with progressively weak demonstrations using in-domain and out-of-domain experts, as well as those trained with and without RoFL.
	
	\paragraph{In-Domain Weak Expert DQfD} We train in-domain reduced and no label experts on the MultiWOZ dataset. The RLE scores 77~F1 on the reduced label test set, while the NLE manages 71~F1 of predicting whether an agent response belongs to a user utterance on the unannotated test set.
	As shown in Table~\ref{tab:baselines} (middle), the scores of DQfD agents with in-domain experts follow a clear trend corresponding to the type of demonstration data. After 2.5 million training steps, the FLE -- with the most informative demonstrations -- clearly outperforms both RLE and NLE methods, while the latter two perform similarly.
	
\begin{figure}
		\centering
		\includegraphics[width=.95\columnwidth]{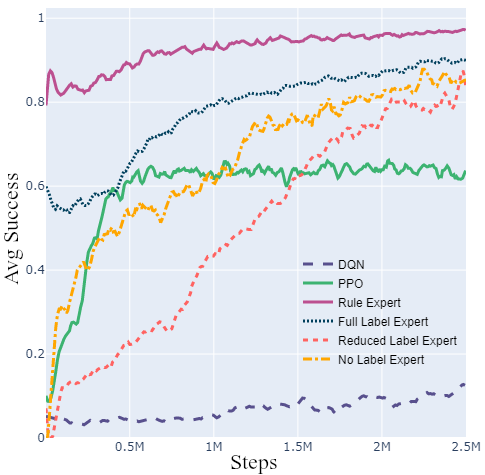}
		\caption{Average Success Rates of our methods trained on \emph{in-domain} data over the course of 2.5 million training steps.}
		\label{fig:in-domain-success}
	\end{figure}

	Figure~\ref{fig:in-domain-success} shows graphs of the average Success Rates of DQN, PPO, and our proposed DQfD agents over the course of training. DQN struggles to find successful dialog strategies, although its Success Rate slowly inclines and seems to gain some traction towards the end of the maximum training steps. To begin with PPO learns rapidly, faster than RLE and NLE, but its Success Rate plateaus in the 60\% range; it seems to learn to end dialogues too early. Both RE and FLE start with performance advantages, due to their high quality expert demonstrations. Over time, RE even approaches the Success Rate of its rule-based expert demonstrator. The FLE consistently outperforms approaches with weaker demonstrations, quickly exceeding the Success Rate of the underlying FFN after an early dip when the agent's exploration parameter $\epsilon$ is relatively high.
	
	The NLE comfortably outperforms the Reduced Label Expert throughout training, with the RLE only overtaking it at the end. We believe that this strong relative performance makes sense if we consider that, during pre-training, the NLE acts according to a more fine-grained action set than the RLE. While the RLE partitions the actions according to their reduced label, these sets are broad and contain many irrelevant responses, whereas the NLE acts randomly according to a smaller, potentially higher-quality, set of actions which have high correspondence scores.
	
	Finally, the graphs in Figure~\ref{fig:in-domain-success} indicate that none of the agents fully converge after the training step limit, although RE and FLE plateau. It is possible that after significantly more steps even DQN would converge to the ceiling performance of the Rule DM -- but all our methods are considerably more sample efficient.
	
	\begin{figure}
		\centering
		\includegraphics[width=.95\columnwidth]{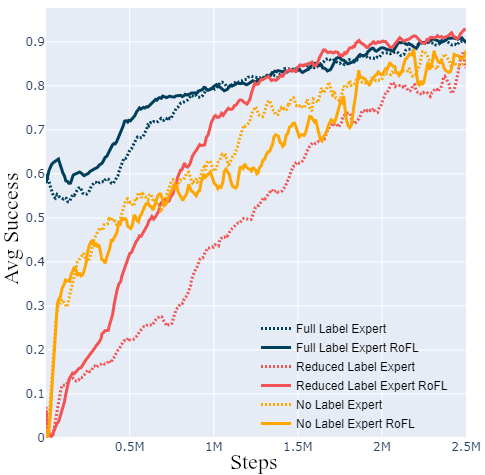}
		\caption{Average Success Rate of RL agents over the course of 2.5 million training steps, with and without RoFL fine-tuning. Agents were evaluated every 2000 steps on 100 evaluation dialogs. Experts were trained on \emph{in-domain} data.}
		\label{fig:in-domain-RoFL-success}
	\end{figure}
	
	\paragraph{RoFL Training} Table~\ref{tab:baselines} (bottom) shows evaluation results of DQfD agents trained with RoFL fine-tuning. All weak experts improve with RoFL, especially the RLE which records an 8\% jump in Success Rate. We also include the performance of the final fine-tuned FFN classifier, whose improvement over its original incarnation (15\% higher Success Rate) demonstrates that fine-tuning helps narrow the domain gap between data and the RL environment. 
	
	In addition to Table~\ref{tab:baselines}, Figure~\ref{fig:in-domain-RoFL-success} shows DM performance over the course of training. RoFL dramatically improves both the performance and convergence rate of the RLE, indicating a domain gap between the reduced label data and the sets of environment actions. RoFL improves the FLE early in training, but this gain tails off after 1 million steps -- possibly due to the relative strength of the expert. The trend for NLE-R is more ambiguous, falling behind its standard DQfD counterpart before catching up to its performance. RoFL seems to lead to the greatest gains when the expert initially struggles.

\begin{figure}[h]
		\centering
		\includegraphics[width=.95\columnwidth]{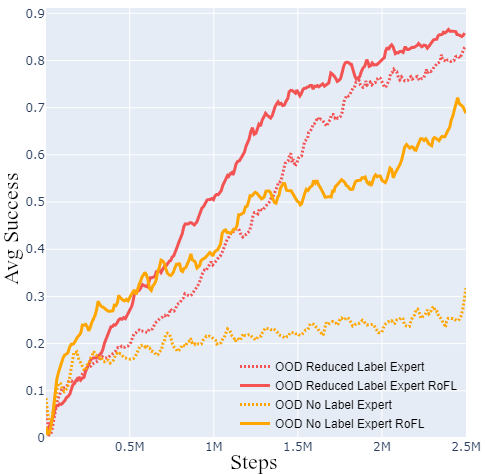}
		\caption{Average Success Rates of Reduced and No Label experts trained on \emph{out-of-domain} data over the course of 2.5 million training steps, with and without RoFL fine-tuning.}
		\label{fig:ood-success}
	\end{figure}

	\paragraph{Out-of-Domain Weak Experts}
	The weakest experts that we evaluate were trained on out-of-domain data. The OOD RLE, trained on Microsoft E2E and Frames, scores 53~F1 on a reduced label MultiWOZ test set, while the OOD NLE, trained on the same datasets, unannotated, only manages 41~F1 on the test set. Results for OOD approaches trained with and without RoFL are shown in Table~\ref{tab:out-domain}, with training graphs in Figure~\ref{fig:ood-success}.
	
	Even without RoFL, the OOD RLE guides the DQfD agent to performance rates comparable to its in-domain counterpart. This indicates that even reduced labels learned on the OOD data  provide the agent with enough clues to correctly satisfy some user goals. With RoFL, the OOD RLE surpasses the Success Rate of the in-domain system, and is only marginally worse than the fine-tuned in-domain expert. This shows that with RoFL we can learn a competitive DM in a challenging multi-domain environment while only using unannotated data from other dialog tasks.
	
	\begin{table}[h]
		\centering
		\setlength{\tabcolsep}{4.5pt}
		\begin{tabular}{l|c|c|c|c}
			\toprule
			& Turns & Inform & Match & Success \\
			\midrule
			RLE & 7.60 & 80.33 & 87.30 & 85.00\\
			NLE & 16.27 & 40.00 & 27.15 & 26.55 \\
			RLE+R & \textbf{6.64} & \textbf{85.00} & \textbf{89.03} & \textbf{91.00}\\
			NLE+R & 9.94 & 70.00 & 62.64 & 68.90\\
			\bottomrule
		\end{tabular}
		\caption{Results of \emph{out-of-domain} weak experts with and without RoFL training, using an agenda-based user-simulator for 1000 evaluation dialogs. Reported scores are average number of Turns, Inform F1, Match Rate, and Success Rate.}
		\label{tab:out-domain}
	\end{table}
	
	RoFL leads to the greatest gain with the OOD NLE. Without fine-tuning, it scores a measly 26\% Success Rate (although it should be noted that this is still higher than DQN), compared to 86\% when the expert is trained on in-domain sentences. This illustrates the clear difference between the language in the in- and out-of-domain data. With RoFL, OOD NLE is able to update its weights to adapt to the language of the environment, outperforming the unaltered expert's Success Rate by 35\%. This improvement holds true throughout training, as shown in Figure~\ref{fig:ood-success}. The graph also shows that OOD NLE+R has not started to converge after 2.5 million training steps; it is likely that with more training it would perform similarly to the in-domain NLE DM.
	
	\section{Conclusions and Future Work}
	In this paper, we have shown that weak demonstrations can be leveraged to learn an accurate dialog manager with Deep Q-Learning from Demonstrations in a challenging multi-domain environment. We established that expert demonstrators can be trained on labeled, reduced-labeled, and unlabeled data and still guide the RL agent by means of their respective auxiliary losses. Evaluation has shown that all experts exceeded the performance of reinforcement and supervised learning baselines, and in some cases even approached the results of a hand-crafted rule-based dialog manager.
	
	Furthermore, we introduced \emph{\textbf{R}einf\textbf{o}rced \textbf{F}ine-tune \textbf{L}earning} (RoFL) a DAgger-inspired extension to DQfD which allows a pre-trained expert to adapt to an RL environment on-the-fly, bridging the domain-gap. Our experiments show that RoFL training is beneficial across different sources of demonstration data, boosting both the rate of convergence and final system performance. It even enables an expert trained on unannotated out-of-domain data to guide an RL dialog manager in a challenging environment.
	
	In future, we want to continue to investigate the possibility of using even weaker demonstrations. Since our No Label Expert is trained on unannotated data, it would be interesting to leverage large and noisy conversational datasets drawn from message boards or movie subtitles, and to see how RoFL training fares with such a significant domain gap between the data and the RL environment.
	
	\section*{Acknowledgements}
	We thank Ignacio Iacobacci and Gerasimos Lampouras for their valuable suggestions.
	
	\bibliography{acl}
	\bibliographystyle{acl_natbib}
	
	\appendix
	
	\section{Model Hyperperameters}
	\label{sec:supplemental}
	Below we list the hyperparameters used for our reinforcement learning agents and the expert models used to generate demonstrations.
	
	\flushleft
	For \emph{No Label Expert} RoFL we treat dialogs with a final reward $r >= th$ as positive, and those with reward $r < th$ as negative examples, and treat the individual user-agent utterance pairs accordingly.
	
	\paragraph{General Hyperparameters}
	Unless otherwise stated in specific expert sections, all of our agents use below hyperparameters, where applicable:
	
	\flushleft
	\begin{tabularx}{\columnwidth}{@{}p{.45\columnwidth}|X}
		\toprule
		Steps & 2,500,000\\
		Pre-training steps $d$ & 2,000 dialogs\\
		$\epsilon$ start value & 0.1\\
		$\epsilon$ end value & 0.01\\
		$\epsilon$ decay rate & Linear over 500,000 steps\\
		Discount factor $\gamma$ & 0.9\\
		Policy net $\pi$ & 1x100d hidden layer, ReLU activation\\
		Learning rate & 0.01\\
		Target network update period $\tau$ & 10,000 steps\\
		L2 reg. weight & $10^{-5}$\\
		Max replay size & 100,000\\
		Prioritized replay $\alpha$ & 0.6\\
		Prioritized replay $\epsilon_p$ & 0.001\\
		Prioritized replay $\epsilon_d$ & 1.0\\
		Prioritized replay $\beta_0$ & 0.4\\
		\bottomrule
	\end{tabularx}
	
	\paragraph{Full Label Expert}
	\begin{tabularx}{\columnwidth}{@{}p{.45\columnwidth}|X}
		\toprule
		Input & 392d binary state\\
		Expert network & 1x150d hidden layer, ReLU, dropout = 0.1\\
		Output & Single-label softmax over 300 actions\\
		\midrule
		\midrule
		Penalty $c$ & 0.8\\
		Reward threshold $th$ &  70\\
		Finetune interval $k$ & 2,000 steps\\
		\bottomrule
	\end{tabularx}
	
	\paragraph{Reduced Label Expert}
	\begin{tabularx}{\columnwidth}{@{}p{.45\columnwidth}|X}
		\toprule
		Input & 300d pre-trained GloVe embeddings\\
		Expert network & 1x128d GRU layer, input dropout = 0.1\\
		Output & Multi-label sigmoid of 3 reduced actions, with threshold = 0.5\\
		\midrule
		\midrule
		Pre-training steps $d$ & 3,000 dialogs\\
		$\epsilon$ start value & 0.2\\
		Penalty $c$ & 1.0\\
		Reward threshold $th$ & 70\\
		Finetune interval $k$ & 15,000 steps\\
		\bottomrule
	\end{tabularx}
	
	\paragraph{No Label Expert}
	\begin{tabularx}{\columnwidth}{@{}p{.45\columnwidth}|X}
		\toprule
		Input & BERT$_{base-uncased}$ embeddings for user and agent utterances, concatenated 1536d\\
		Expert network & 6 linear layers, 512, 256, 128, 64, 32, 16 dims, SELU activations\\
		Output & Binary output of response appropriateness with threshold $\rho$ = 0.9\\
		\midrule
		\midrule
		Penalty $c$ & 0.8\\
		Reward threshold $th$ & 70\\
		Finetune interval $k$ & 15,000 steps\\
		\bottomrule
	\end{tabularx}
\end{document}